\pdfoutput=1

\documentclass[11pt]{article}

\usepackage[final]{acl}

\usepackage{times}
\usepackage{latexsym}

\usepackage[T1]{fontenc}

\usepackage[utf8]{inputenc}
\usepackage{amsmath}
\usepackage{amssymb}

\usepackage{microtype}

\usepackage{inconsolata}

\usepackage{graphicx}
\usepackage{amsmath}
\usepackage{subcaption}
\usepackage{float}
\usepackage{multirow}
\usepackage{booktabs}
\usepackage{colortbl}
\usepackage{titletoc}

\title{Sensitivity-LoRA : Low-Load Sensitivity-Based Fine-Tuning for Large Language Models}

\author{\bf Hao Zhang$^{1}$\thanks{These authors contribute equally to this work.}, Bo Huang$^{2}$\footnotemark[1], Zhenjia Li$^{3}$\footnotemark[1], Xi Xiao$^4$,  Huiyi Leong$^5$, Zumeng Zhang$^6$, \\ \bf Xinwei Long$^7$, Tianyang Wang$^4$, Hao Xu$^{1}$\thanks{Corresponding author.}\\
\normalsize{}$^1$Harvard University,
\normalsize{}$^2$Xi'an University of Science and Technology \\
\normalsize{}$^3$University of Chinese Academy of Sciences, 
\normalsize{}$^4$University of Alabama at Birmingham \\ 
\normalsize{}$^5$University of Chicago,
\normalsize{}$^6$Yunnan University,
\normalsize{}$^7$Tsinghua University \\ 
\normalsize{}zh.cs.star@outlook.com, haxu@bwh.harvard.edu}

\begin{document}
\maketitle
\begin{abstract}
Large Language Models (LLMs) have transformed both everyday life and scientific research.  However, adapting LLMs from general-purpose models to specialized tasks remains challenging, particularly in resource-constrained environments. Low-Rank Adaptation (LoRA), a prominent method within Parameter-Efficient Fine-Tuning (PEFT), has emerged as a promising approach to LLMs by approximating model weight updates using low-rank decomposition. However, LoRA is limited by its uniform rank \( r \) allocation to each incremental matrix, and existing rank allocation techniques aimed at addressing this issue remain computationally inefficient, complex, and unstable, hindering practical applications. 
To address these limitations, we propose \textbf{Sensitivity-LoRA}, an efficient fine-tuning method that dynamically allocates ranks to weight matrices based on both their global and local sensitivities. It leverages the second-order derivatives (Hessian Matrix) of the loss function to effectively capture weight sensitivity, enabling optimal rank allocation with minimal computational overhead. 
Our experimental results have demonstrated robust effectiveness, efficiency and stability of Sensitivity-LoRA across diverse tasks and benchmarks.
\end{abstract}

\section{Introduction}
Large language models (LLMs) have become transformative tools across a wide spectrum of tasks and applications \cite{ding2022, qin2023,zhu2023,zhu2023a,li2023,zhang2023,huang2023,wang2023,ma2025cad,wu2025sugar,qi2025mediaug,qi2025medconv,luo2025pathohr,zhang2025towards,zhang2025enhancingllmefficiencytargeted,Leo2024AgentNet,Zhang2025TimeLLaMA}. Despite these advancements, fine-tuning remains a critical technique for effectively adapting LLMs from general-purpose models to specialized applications, especially in resource-constrained environments.
However, full-parameter fine-tuning can be prohibitively resource-intensive, requiring significant computational power and GPU capacity. To address this limitation, the research community introduced parameter-efficient fine-tuning (PEFT) \cite{lester2021,li2021,zaken2022,hu2022,xiao2025visualinstanceawareprompttuning,xiao2025focusfusedobservationchannels}, which aims to balance accuracy and efficiency by selectively updating a subset of model parameters.

LoRA \cite{hu2022lora}, a prominent PEFT method, approximates model weight updates using low-rank decomposition, leveraging the low intrinsic dimension of over-parameterized models \cite{li2018measuring,aghajanyan2020intrinsic}.
During training, the update of the weight matrix \( (\Delta W) \) can be approximated as the product of two smaller matrices \( B \) and \( A \), expressed as:
\begin{equation}
\Delta W \approx B \cdot A
\end{equation}

where \(\Delta W \in \mathbb{R}^{d_1 \times d_2}\), \(A \in \mathbb{R}^{ r \times d_2}\) and \(B \in \mathbb{R}^{ d_1 \times r}\) with \(r \ll \{d_1, d_2\}\). Thus, it approximates the update of the weight matrix with fewer parameters. However, the full potential of LoRA remains constrained by its inherent design limitations. Specifically, it assumes a uniform rank \textit{r} for each incremental matrix, not accounting for the varying significance of weight matrices across different modules and layers \cite{hu2023structure,zhang2023increlora}. 
\begin{figure*}[t]
    \centering
    \includegraphics[width=0.98\linewidth]{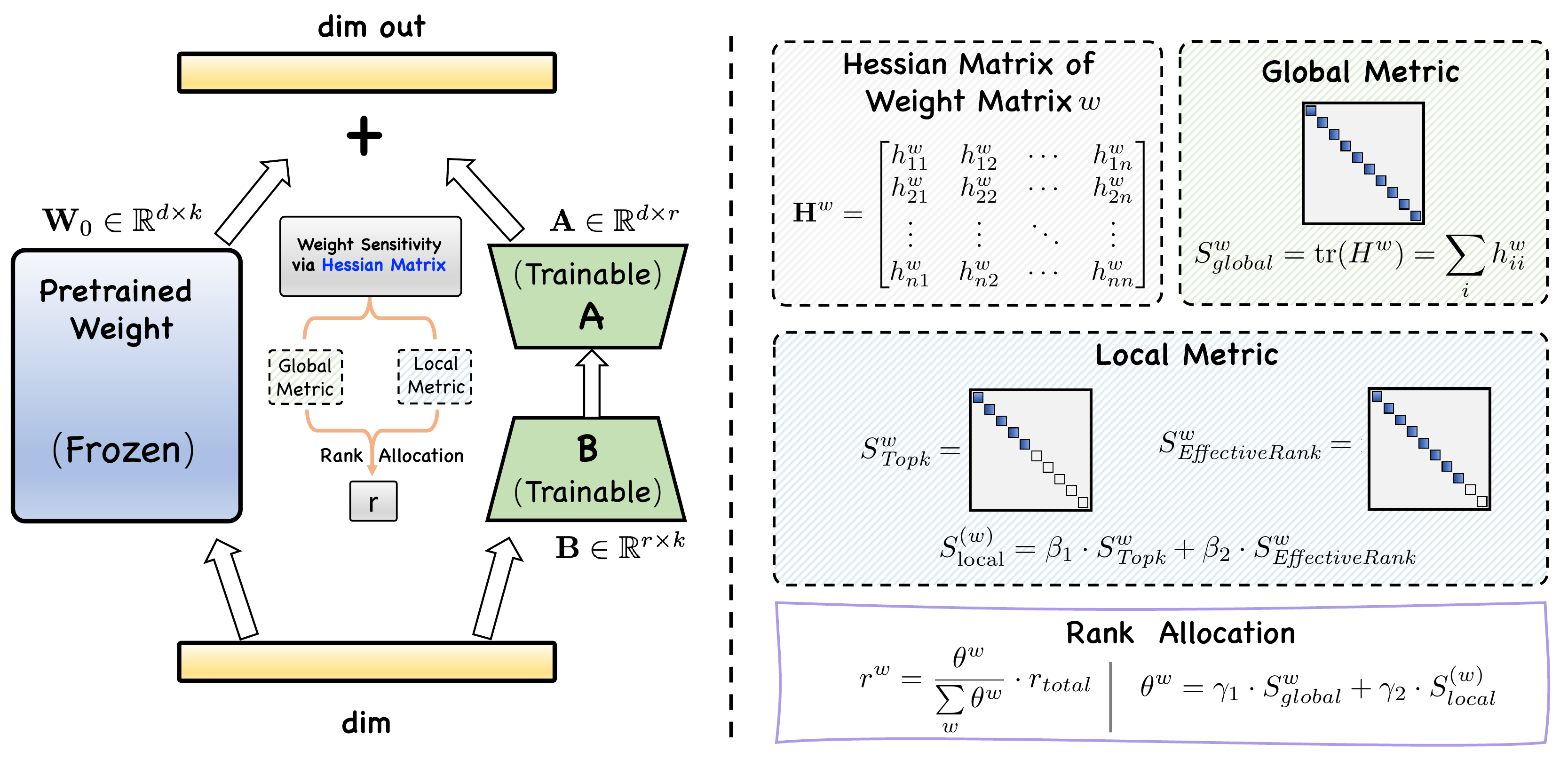}
    \caption{Pipeline of the Sensitivity-LoRA Method: \textbf{Step 1 - Sensitivity detection} via Hessian-based metrics, including global and local sensitivity measures. ($h_{ij}^{w} = \frac{\partial^2 E^{w}}{\partial w_i \partial w_j}, h_{ii} = \frac{\partial^2 E^{w}}{\partial w_i^2}$, where \(E^{w}\) denotes the change in loss function regarding weight matrix \(w\); \( \text{tr}(H^{w}) \) denotes the trace of \(H^{w}\).) \textbf{Step 2 - Dynamic rank allocation} based on global and local sensitivity. (\(r^{w}\) denotes the allocated rank of weight matrix \(w\), \(r_{total}\) denotes the total rank of all weight matrices in the model.)}
    \label{fig:fig1}
\end{figure*}

To address this limitation, dynamic rank allocation has emerged as a key solution by allocating the rank \textit{r} to each different module or layer according to its specific requirements. Existing methods achieve this through three main approaches: singular value decomposition (SVD), single-rank decomposition (SRD), and rank sampling.
SVD-based methods \cite{zhang2023adalora,hu2023structure,zhang2023increlora} decompose matrix BA into an SVD form and selectively truncate the singular values in order to allocate the matrix rank. However, this process is computationally expensive and requires additional memory to store singular values and vectors.
SRD-based methods \cite{mao2024dora,zhang2024autolora,liu2024alora} decompose matrix BA into single-rank components and allocate the ranks by selecting the proper components. However, optimizing single-rank components and the pruning process can increase computational complexity, potentially offsetting efficiency gains. 
Rank sampling-based methods \cite{valipour2022dylora} allocate ranks directly by random sampling. However, the randomness introduced by sampling could increase training instability.

In order to design a dynamic rank allocation method that introduces extremely low overhead and ensures stability, we propose Sensitivity-LoRA, which can rapidly allocate rank to the weight matrix based on the sensitivity of the parameters, without incurring a significant computational load. Specifically, we utilize the second derivatives of the loss function with respect to the parameters (Hessian matrix) to ascertain the sensitivity of each parameter within the weight matrix. To comprehensively evaluate the sensitivity of the parameter matrix, we employ metrics such as the trace of the Hessian matrix, \textit{Topk} and \textit{Effective Rank} to measure its global and local sensitivities respectively. By integrating various metrics, we determine the rank allocation weights corresponding to the weight matrices to achieve rank allocation. The efficiency, stability, and generality of our approach have been validated through extensive experiments on various tasks, such as sentiment analysis, natural language inference, question answering, and text generation. 


In summary, the main contributions of our paper are listed as follows:

\begin{itemize}
\item We design a dynamic rank allocation method that introduces minimal overhead and ensures stability.
\item We introduce the second derivatives of the loss function with respect to the weight matrix to measure their sensitivity.
\item We achieve rank allocation by taking into account both the global and local sensitivity of the weight matrix.
\item Extensive experiments demonstrate the effectiveness, stability, and efficiency of our method.
\end{itemize}

\section{Related Work}
Existing PEFT approaches can be classified into four main types in terms of memory efficiency, storage efficiency, and inference overhead, as follows:
\subsection{Additive PEFT}
Additive PEFT introduces lightweight modules into the model architecture, such as adapters and soft prompts, while keeping the pre-trained backbone frozen. Adapters add small networks with down-projection and up-projection matrices, enabling task-specific learning with minimal parameter updates \cite{houlsby2019, lester2021}. Soft prompts prepend learnable embeddings to the input sequence, allowing fine-tuning by modifying input activations only \cite{li2021, zaken2022}. These methods typically require updating less than 1\% of the total parameters, significantly reducing computation and memory costs, making them ideal for resource-constrained environments \cite{hu2022}.
\subsection{Selective PEFT}
Selective PEFT fine-tunes a subset of the existing parameters in a pre-trained model, rather than adding new modules. It employs binary masks to identify and update only the most important parameters while keeping the majority frozen. Techniques like Diff pruning and FishMask leverage Fisher information or parameter sensitivity analysis to select critical parameters for fine-tuning \cite{zaken2022, li2021}. This approach avoids increasing model complexity and is particularly suited for scenarios where only a small fraction of the model contributes significantly to performance.
\subsection{Reparameterized PEFT}
Reparameterized PEFT utilizes low-rank parameterization techniques to represent model weights in a reduced form during training. LoRA (Low-Rank Adaptation) is a prominent example, introducing low-rank matrices to fine-tune specific weights while maintaining high inference efficiency \cite{hu2022}. Other methods, such as Compacter, use the Kronecker product for parameter reparameterization, further reducing memory requirements and computational costs \cite{houlsby2019}. Reparameterized PEFT is highly effective for large-scale models where resource constraints are critical.
\subsection{Hybrid PEFT}
Hybrid PEFT combines the strengths of Additive, Selective, and Reparameterized PEFT methods into a unified framework. For example, UniPELT integrates LoRA, adapters, and soft prompts, allowing dynamic selection of the most suitable module for specific tasks through gating mechanisms \cite{zaken2022}. This hybrid approach enhances adaptability and task performance by leveraging the complementary advantages of different PEFT strategies \cite{li2021, hu2022}.

\section{Methodology}
\label{Methodology}
In this section, we firstly introduce the concept of weight sensitivity with a formal definition of global and local sensitivity metrics of weight matrices. Next, we propose effective allocation strategies to optimize the dynamic rank allocation process based on these sensitivity metrics. The pipeline of our method is presented in Figure \ref{fig:fig1}.
\subsection{Weight Sensitivity}
Consider a neural network whose dynamics is driven by a collection of parameters \(w\) and a loss function \(E\), which guides its learning dynamics. 
When a small perturbation \( \delta w \) is introduced to the parameters, the resulting change in the loss function can be expressed using a Taylor series expansion up to the second-order term, with higher-order terms captured by $O(\|\delta w\|^3)$ as follows:
\begin{align}
E(w + \delta w) &= E(w) + g^T \delta w + \frac{1}{2} \delta w^T H \delta w \notag \\
&\quad + O(\|\delta w\|^3)
\end{align}
where \(g\) denotes the gradient vector of the loss function \(E\) with respect to the parameters \(w\), indicating the rate of change of the loss function in the direction of each parameter. \(H\) represents the Hessian matrix of the loss function \(E\), which is a matrix of second-order partial derivatives and contains information about the curvature of the loss function at the current parameter point. 

The change in the loss function $\Delta E$ can be represented by the following expression:
\begin{equation}
    \Delta E =  g^\top \delta\mathbf{w} + \frac{1}{2} \delta\mathbf{w}^\top H \delta\mathbf{w} + O(\|\delta_w\|^3)
\end{equation}
By expanding the components of $\Delta E$, we have:
\begin{align}
    \Delta E &= \sum_i g_i \delta w_i + \frac{1}{2} \sum_{i,j} h_{ij} \delta w_i \delta w_j \notag \\
    &\quad + O(\|\delta_w\|^3)
\end{align}
where $g_i$ and $h_{ij}$ are the gradient and Hessian elements, respectively. 

For a well-trained neural network, when the parameter \(w\) is located at a local minimum of the loss function, the gradient \(g\) becomes zero. Then, the above
equation can be simplified to
\begin{equation}
    \Delta E = \frac{1}{2} \sum_{i,j} h_{ij} \delta w_i \delta w_j + O(\|\delta_w\|^3)
\end{equation}
Additionally, several studies have demonstrated that the Hessian matrix \(H\) tends to be diagonally dominant, suggesting that the interactions between different parameters can be largely disregarded \cite{lecun1989optimal,dong2020hawq,frantar2022gptq}. Then, the above equation can be simplified to
\begin{equation}
    \Delta E \approx \frac{1}{2} \sum_i h_{ii} \delta_{w_i}^2 + O(\|\delta_w\|^3)
\end{equation}
Given that the perturbation in the weights (\(\delta \mathbf{w}\)) is sufficiently small, the higher-order term becomes negligible compared to the quadratic term, and therefore can be disregarded. Consequently, the above formula can be further simplified to
\begin{equation}
    \Delta E \approx \frac{1}{2} \sum_i h_{ii} \delta_{w_i}^2
\end{equation}
Consequently, the diagonal elements of the Hessian matrix serve as a reliable indicator of weight sensitivity. 
\subsection{Rank Allocation Metric}
\subsubsection{
Global Metric}
The global sensitivity measurement aims to evaluate the overall impact of an entire parameter (or weight) matrix on model output. It quantifies how variations in this weight matrix affect the loss function. To capture this dynamics, the Hessian matrix, which consists of the second-order partial derivatives of the loss function with respect to the weight matrix, is used. Given that the Hessian matrix tends to be diagonal-dominant at the minimum, its trace can serve as an effective global sensitivity indicator. Formally, the global sensitivity \( S_{global}^{w}\) of weight matrix \(w\) can be defined as:
\begin{equation}
     S_{global}^{w} = \text{tr}(H^{w}) = \sum_{i} h_{ii}^{w} 
\end{equation}
where \( h_{ii}^{w} \) is the \( i \)-th diagonal element of the Hessian matrix \( H^{w} \), and \( \text{tr}(H^{w}) \) denotes the trace of \(H^{w}\).
Since the diagonal elements reflect the impact of individual parameter changes on the loss function, a larger trace value indicates that the model is more sensitive to its changes. This suggests more parameters make significant contributions to the changes in the loss function, emphasizing their role in model performance.

\subsubsection{Local Metric}

While certain weight matrices might have low overall sensitivities, specific weight elements within these matrices can still have high sensitivity, significantly impacting model performance. As such, it is essential to account for local sensitivity to capture finer-grained variations in parameter influence on the loss function. To address this, we introduce two metrics: \textit{Topk} and \textit{Effective Rank}.



The \textit{Topk} metric approximates local sensitivity of a weight matrix by averaging its largest \textit{k} diagonal elements of Hessian matrix, based on the assumption that most of the matrix’s energy or sensitivity is concentrated in these large values. 
By focusing on \textit{Top k} diagonal elements, the \textit{Topk} metric can guide us to prioritize these critical weights during weight pruning or optimization processes. It reduces computational complexity while preserving the most impactful weights for model performance.
The computation formula for the \textit{Topk} metric of weight matrix \(w\) is as follows:
\begin{equation}
    S_\textit{{Topk}}^{w} = \frac{1}{k} \sum_{i=1}^{k} \lambda_i^{w}
\end{equation}
where \( \lambda_i^{w} \) represents the diagonal elements of Hessian matrix \( H^{w} \) sorted in descending order, and \( k \) denotes the number of diagonal elements selected.


The \textit{Effective Rank} metric determines the minimum rank that captures most of the energy of a weight matrix based on the cumulative contribution of the diagonal elements of its Hessian matrix. By establishing a threshold for the cumulative contribution rate (such as 0.9 or 0.95), the \textit{Effective Rank} metric identifies the minimum number of eigenvalues needed to achieve this threshold, thereby appropriately ranking the weight matrix. The key benefit of this metric is ensuring the stability of the rank allocation process. The formula for \textit{Effective Rank} of weight matrix \(w\) is as follows:
\begin{equation}
S_{\textit{EffectiveRank}}^{w} = \min \left\{ k \mid \frac{\sum_{j=1}^{k} \lambda_j^{w} }{\sum_{j=1}^{m} \lambda_j^{w} } \geq \alpha \right\}
\end{equation}
where \( \lambda_j^{w}  \) is the \( j \)-th diagonal element of \( H^{w} \) in non-increasing order, \( m \) is the total number of diagonal elements, and \( k \) is the minimum number of diagonal elements required for the cumulative contribution rate to reach the threshold \( \alpha \).

To ensure the effectiveness and stability, we integrate \textit{Topk} and \textit{Effective Rank} metrics together to define the local sensitivity metric \( S_{local}^{w}  \) of weight matrix \(w\) as follows: 
\begin{equation}
\beta_1 = \frac{\sigma^{S_{T}}}{\left(\mu^{S_{T}}\right)^2} \hspace{0.25cm} \beta_2 = \frac{\sigma^{S_{E}}}{\left(\mu^{S_{E}}\right)^2}
\end{equation}
\begin{equation}
S_{local}^{w} = \beta_1 \cdot S_\textit{{Topk}}^{w} + \beta_2 \cdot S_{\textit{EffectiveRank}}^{w}
\end{equation}
where \(\sigma^{S_{T}}\) and \(\sigma^{S_{E}}\) represent the standard deviations of the \textit{Topk} and \textit{Effective Rank} metrics for all weights, while \(\mu^{S_{T}}\) and \(\mu^{S_{E}}\) denote the corresponding mean values. We utilize the standard deviation of metrics to design allocation weights. The larger the standard deviation of a metric, the more widely its values are distributed, which imply a greater amount of information contained within that metric. The squared average values represent the normalization of the metric scale and the standard deviation. The effectiveness of this design is demonstrated through experiments.

\subsection{Rank Allocation Strategy}\label{3.3}
Taking into account both global and local metrics, we define a refined rank allocation strategy to determine the rank allocation weights \(\theta^{w} \) of weight matrix \(w\) by integrating global and local sensitivities:
\begin{equation}
\gamma_1 = \frac{\sigma^{S_{g}}}{\left(\mu^{S_{g}}\right)^2} \hspace{0.25cm} \gamma_2 = \frac{\sigma^{S_{l}}}{\left(\mu^{S_{l}}\right)^2}
\end{equation}
\begin{equation}
\theta^{w}  = \gamma_1 \cdot S_{global}^{w}  + \gamma_2 \cdot S_{local}^{w} 
\end{equation}
where \(\sigma^{S_{g}}\) and \(\sigma^{S_{l}}\) represent the standard deviations of the global and local metrics for all weights, while \(\mu^{S_{g}}\) and \(\mu^{S_{l}}\) denote the corresponding mean values. The reason for this design is mentioned in the preceding text. Hence, we can derive the formula for rank allocation as follows:
\begin{equation}
    r^{w} = \frac{\theta^{w}}{\sum\limits_{w} \theta^{w}} \cdot r_{total}
\end{equation}
where \(r^{w}\) denotes the rank allocated to weight matrix \(w\), and \(r_{total}\) represents the total rank of all weight matrices in the model.

\section{Experiments}
\label{Experiments}
\subsection{Experimental Setup}
\textbf{Models and Benchmarks. }We evaluate the performance of our method across diverse NLG (Natural Language Generation) and NLU (Natural Language Understanding) tasks. For the NLU tasks, we select RoBERTa-base \cite{liu2019roberta} as the base model and evaluate its performance on various subtasks of the GLUE (General Language Understanding Evaluation) benchmark \cite{wang2018glue}: MNLI \cite{williams2017mnli}, SST-2 \cite{socher2013sst2}, MRPC \cite{dolan2005mrpc}, CoLA \cite{warstadt2019cola}, QNLI \cite{rajpurkar2018qnli}, QQP \footnote{https://quoradata.quora.com/First-Quora-Dataset-Release-Question-Pairs}, RTE \cite{wang2018glue} and STS-B \cite{cer2017stsb}. 
For the NLG tasks, we conduct experiments using two large language models, Qwen2.5-7B \cite{yang2024qwen2} and LLaMA3.1-8B \cite{grattafiori2024llama}, and evaluate their performance on two representative NLG datasets: Magpie-Pro \cite{xu2024magpie} and OpenPlatypus \cite{lee2023platypus}. We also visualize the global and local rank allocation results for each layer of GPT-2 Large \cite{radford2019language} and RoBERTa-base \cite{liu2019roberta}.

\textbf{Evaluation Metrics. }We report a comprehensive set of standard evaluation metrics. For NLG tasks, we utilize BLEU \cite{papineni2002bleu} and ROUGE \cite{lin2004rouge} to assess the quality of generated text. For NLU tasks, we employ the Matthew’s correlation coefficient for the CoLA task, the Combined Score for STS-B, and accuracy for the remaining NLU tasks. 

\textbf{Baselines. }We adopt several representative methods, including  HAdapter \cite{houlsby2019HAdapter}, PAdapter \cite{pfeiffer2020PAdapter}, LoRA \cite{hu2022lora} with uniform rank allocation, AdaLoRA \cite{zhang2023adalora} and DyLoRA \cite{valipour2022dylora}, as our baselines. More details can be found in the Appendix \ref{Baselines}.

In addition, more implementation details can be found in the Appendix \ref{sec:detailed_exp_setup}.


\begin{table*}[htbp]
\small
\centering
\begin{tabular}{cccccccccc}
\toprule
\textbf{Method} & \textbf{MNLI} & \textbf{SST-2} & \textbf{MRPC} & \textbf{CoLA} & \textbf{QNLI} & \textbf{QQP} & \textbf{RTE} & \textbf{STS-B} & \textbf{Avg.} \\ 
\midrule
HAdapter & 86.76          & 94.03          & 87.01          & 57.84          & 93.19          & 90.42          & 78.75          & 90.91          & 84.86          \\
PAdapter & 86.95          & 94.11          & 86.54          & 57.95          & 93.37          & 90.55          & 79.42          & 90.97          & 84.98          \\
LoRA     & 87.26          & 93.46          & 87.08          & 58.83          & 92.95          & 90.50          & 79.39          & 91.03          & 85.06          \\
AdaLoRA  & 87.32          & 93.57          & 87.28          & 59.00          & 93.08          & 90.62          & 79.56          & 91.21          & 85.20          \\
DyLoRA & 87.24 & 93.65 & 87.28 & 58.98 & 93.00 & 90.57 & 79.59 & 91.17 & 85.19 \\
\rowcolor[gray]{0.8} 
Sensitivity-LoRA (ours)     & \textbf{87.58} & \textbf{94.59} & \textbf{87.73} & \textbf{60.20} & \textbf{93.62} & \textbf{90.74} & \textbf{81.81} & \textbf{91.27} & \textbf{85.94} \\ 
\bottomrule
\end{tabular}
\vspace{-5pt}
\caption{Performance comparison between baseline methods and the proposed approach on the GLUE benchmark using the RoBERTa-base model. Higher values indicate better performance across all tasks. Bolded values denote the best performance in each task.}
\label{tab:NLU}
\end{table*}



\subsection{Main Results}
We evaluate the effectiveness of Sensitivity-LoRA on NLU tasks by finetuning the RoBERTa-base model across the tasks in the GLUE benchmark. As shown in Table \ref{tab:NLU}, Sensitivity-LoRA demonstrates outstanding performance in a variety of natural language understanding tasks. Specifically, our method achieves the highest average score of 85.94, outperforming all baselines. Sensitivity-LoRA leverages the second order derivatives of the loss function to extract weight wise importance metrics, incorporating both local and global sensitivity. Based on these metrics, it dynamically determines the optimal rank allocation, thereby achieving exceptional performance.



To further assess the effectiveness of our method on NLG tasks, we compare Sensitivity-LoRA against baselines on two diverse datasets: Magpie-Pro and OpenPlatypus, utilizing Qwen2.5-7B and LLaMA3.1-8B. As shown in Table \ref{tab:NLG}, our method consistently outperforms all baselines on evaluation metrics, including BLEU-4, ROUGE-1, and ROUGE-L. On Qwen2.5-7B, our method achieves the highest average score of 37.98, significantly outperforming others. On LLaMA3.1-8B, it further demonstrates its advantage by attaining an average score of 49.57, surpassing AdaLoRA (48.80), LoRA (48.37), and other adapter based methods. Notably, our method achieves substantial gains in BLEU-4 (71.25) and ROUGE-1 (57.35) on the Magpie-Pro dataset and leads across all three metrics on OpenPlatypus. These results highlight the superior generalization and effectiveness of our sensitivity aware finetuning strategy across various models and generation scenarios.

\begin{figure}[]
    \centering
    \includegraphics[width=0.9\linewidth]{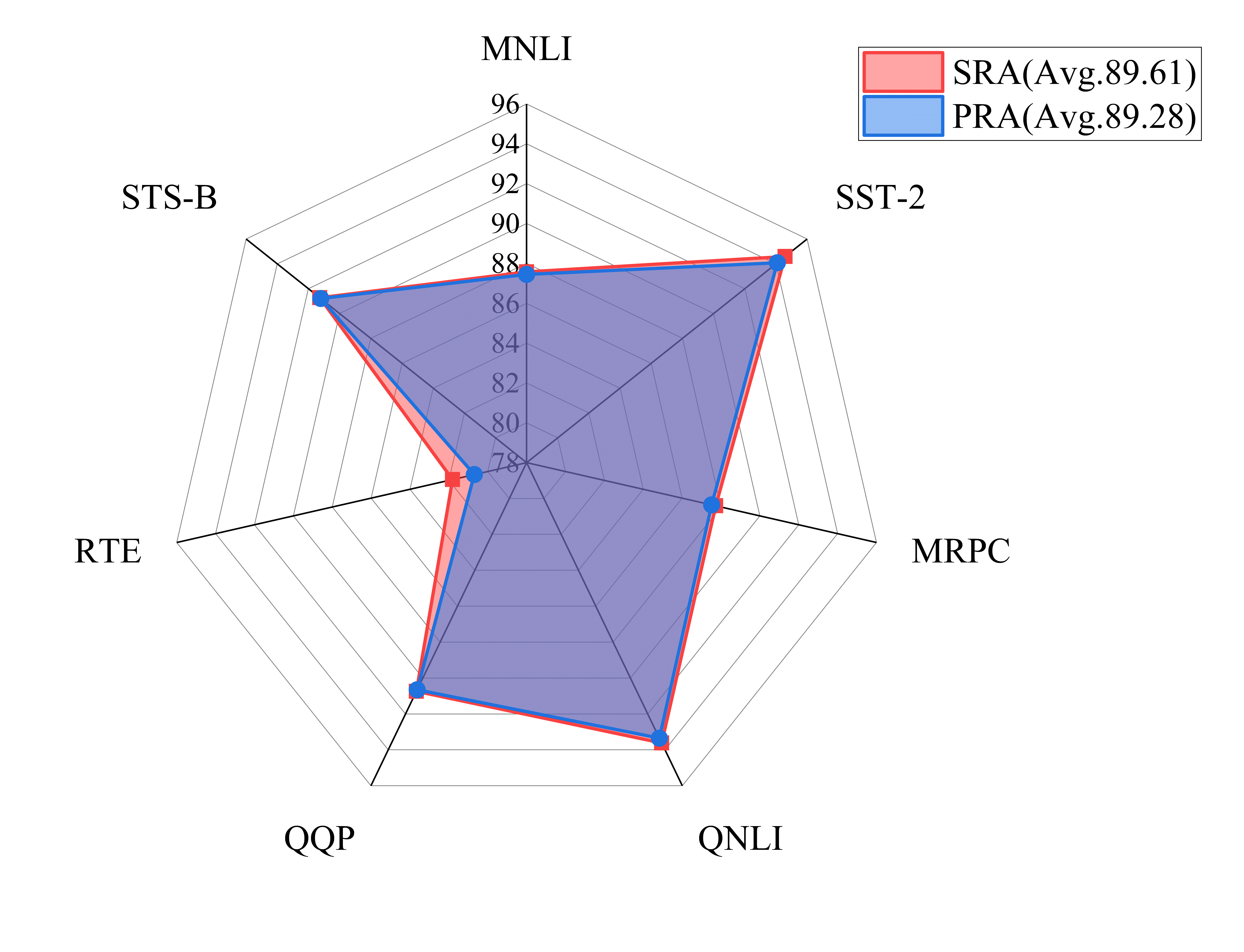}
    \vspace{-15pt}
    \caption{Comparison of the evaluation results for RoBERTa-base finetuned on several datasets from the GLUE benchmark, using both PRA and SRA rank allocation methods.}
    \label{fig:rank_assignment}
\end{figure}

\begin{table*}[htbp]
\centering
\small
\setlength{\tabcolsep}{3.5pt}
\begin{tabular}{ccccccccc}
\toprule
\multirow{2}{*}{\textbf{Model}} & \multirow{2}{*}{\textbf{Method}} & \multicolumn{3}{c}{\textbf{Magpie-Pro}} & \multicolumn{3}{c}{\textbf{OpenPlatypus}} & \multirow{2}{*}{\textbf{Avg.}} \\ 
\cmidrule(lr){3-5} \cmidrule(lr){6-8}
 &  & 
 \textbf{BLEU-4} & \textbf{ROUGE-1} & \textbf{ROUGE-L} & \textbf{BLEU-4} & \textbf{ROUGE-1} & \textbf{ROUGE-L} &  \\ 
\midrule

\multirow{5}{*}{Qwen2.5-7B} & HAdapter & 54.71 & 49.11 & 32.42 & 19.39 & 43.95 & 22.51 & 37.01 \\
 & PAdapter & 54.83 & 49.15 & 32.24 & 19.42 & 44.03 & 22.54 & 37.04 \\
 & LoRA     & 55.03 & 48.82 & 32.42 & 19.72 & 43.83 & 22.53 & 37.06 \\
 & AdaLoRA  & 55.66 & 49.13 & 32.75 & 19.87 & 44.24 & 22.67 & 37.39 \\
 & DyLoRA & 55.59 & 49.21 & 32.82 & 19.86 & 44.18 & 22.59 & 37.37 \\
 & \cellcolor{gray!20}Sensitivity-LoRA (ours) & \cellcolor{gray!20}\textbf{56.31} & \cellcolor{gray!20}\textbf{50.04} & \cellcolor{gray!20}\textbf{33.57} & \cellcolor{gray!20}\textbf{20.13} & \cellcolor{gray!20}\textbf{44.77} & \cellcolor{gray!20}\textbf{23.07} & \cellcolor{gray!20}\textbf{37.98} \\

\midrule

\multirow{5}{*}{LLaMA3.1-8B} & HAdapter & 69.28 & 56.23 & 41.08 & 34.73 & 52.31 & 35.61 & 48.20 \\
 & PAdapter & 69.30 & 55.05 & 41.97 & 34.66 & 51.47 & 36.01 & 48.07 \\
 & LoRA     & 69.67 & 55.89 & 41.78 & 34.64 & 52.35 & 35.92 & 48.37 \\
 & AdaLoRA  & 70.40 & 56.31 & 42.17 & 34.86 & 52.90 & 36.15 & 48.80 \\
 & DyLoRA & 70.36 & 56.26 & 42.16 & 34.89 & 52.80 & 36.20 & 48.78 \\
 & \cellcolor{gray!20}Sensitivity-LoRA (ours) & \cellcolor{gray!20}\textbf{71.25} & \cellcolor{gray!20}\textbf{57.35} & \cellcolor{gray!20}\textbf{43.02} & \cellcolor{gray!20}\textbf{35.30} & \cellcolor{gray!20}\textbf{53.79} & \cellcolor{gray!20}\textbf{36.69} & \cellcolor{gray!20}\textbf{49.57} \\

\bottomrule
\end{tabular}
\vspace{-5pt}
\caption{Evaluation results on NLG tasks using Qwen2.5-7B and LLaMA3.1-8B as backbone models. We compare Sensitivity-LoRA with other PEFT baselines on two representative datasets, Magpie-Pro and OpenPlatypus. Metrics reported include BLEU-4, ROUGE-1, and ROUGE-L.}
\label{tab:NLG}
\end{table*}

\subsection{Ablation Study}
In this section, we present a detailed set of ablation studies to thoroughly evaluate the effectiveness of each component of our method. The evaluation results are summarized in Table~\ref{tab:ablation}, where we finetune the RoBERTa-base model on the GLUE benchmark using three different strategies: the proposed global metric $S_g$, the local metric $S_l$, and their combination. Our findings indicate that both $S_g$-LoRA and $S_l$-LoRA significantly outperform the vanilla LoRA baseline, which employs a uniform rank allocation strategy. This clearly demonstrates that incorporating either global or local sensitivity information can lead to more informed and effective rank assignments. Furthermore, our full method, which integrates both global and local metrics, achieves the highest average score of 85.94. This result underscores the complementary nature of the two types of sensitivity and highlights the benefits of combining both perspectives to guide finetuning. Overall, these results validate the effectiveness of our sensitivity aware rank allocation mechanism and provide strong evidence for the advantages of leveraging both global and local sensitivity information in optimizing model performance.

\begin{table*}[htbp]
\centering
\small
\begin{tabular}{lccccccccc}
\toprule
\textbf{Method} & \textbf{MNLI} & \textbf{SST-2} & \textbf{MRPC} & \textbf{CoLA} & \textbf{QNLI} & \textbf{QQP} & \textbf{RTE} & \textbf{STS-B} & \textbf{Avg.} \\ 
\midrule

LoRA        & 87.26 & 93.46 & 87.08 & 58.83 & 92.95 & 90.50 & 79.39 & 91.03 & 85.06 \\

$S_g$-LoRA  & 87.45 & 94.08 & 87.51 & 59.60 & 93.35 & 90.68 & 80.19 & 91.24 & 85.51 \\

$S_l$-LoRA  & 87.41 & 94.05 & 87.49 & 59.60 & 93.33 & 90.64 & 80.19 & 91.27 & 85.50 \\

\rowcolor[gray]{0.8}
Ours        & \textbf{87.58} & \textbf{94.59} & \textbf{87.73} & \textbf{60.20} & \textbf{93.62} & \textbf{90.74} & \textbf{81.81} & \textbf{91.27} & \textbf{85.94} \\ 
\bottomrule
\end{tabular}
\vspace{-5pt}
\caption{Ablation study on the GLUE benchmark using the RoBERTa-base model. We compare the performance of different rank allocation strategies: the uniform baseline (LoRA), global sensitivity based allocation ($S_g$-LoRA), local sensitivity based allocation ($S_l$-LoRA), and the proposed combined method (Ours).}
\label{tab:ablation}
\end{table*}

\subsection{Comparison of Rank Allocation Methods}\label{4.3}
In this section, we compare two rank allocation methods for model weights, based on global and local sensitivity metrics. The Progressive Rank Allocation (PRA) method first sorts the metrics in descending order, subsequently allocating ranks progressively within a specified range. Weights with higher sensitivity are allocated higher ranks. For example, assume there are 6 weights sorted by sensitivity. The average number of \(r\) allocated to each matrix is 5, and there are 3 categories in total. The allocation of \(r\) for the weights is 6, 6, 5, 5, 4, and 4, respectively. The Scaled Rank Allocation (SRA) method (mentioned in Section \ref{3.3}) allocates ranks according to the proportion of each weight's metric relative to the model's total metrics. To visually compare the effectiveness of these two allocation methods, we apply both strategies to the sensitivity metrics and subsequently combine them using the corresponding rank allocation strategy. We then finetune RoBERTa-base on some datasets from the GLUE benchmark. As illustrated in Figure \ref{fig:rank_assignment}, SRA consistently outperforms PRA on various datasets, demonstrating its robustness and superior adaptability. This consistent improvement across datasets suggests that the SRA enables more effective allocation decisions, leading to better overall results. Consequently, we adapt the SRA method for rank allocation in this paper.

\begin{figure*}[t]
\centering
  \includegraphics[width=0.9\linewidth]{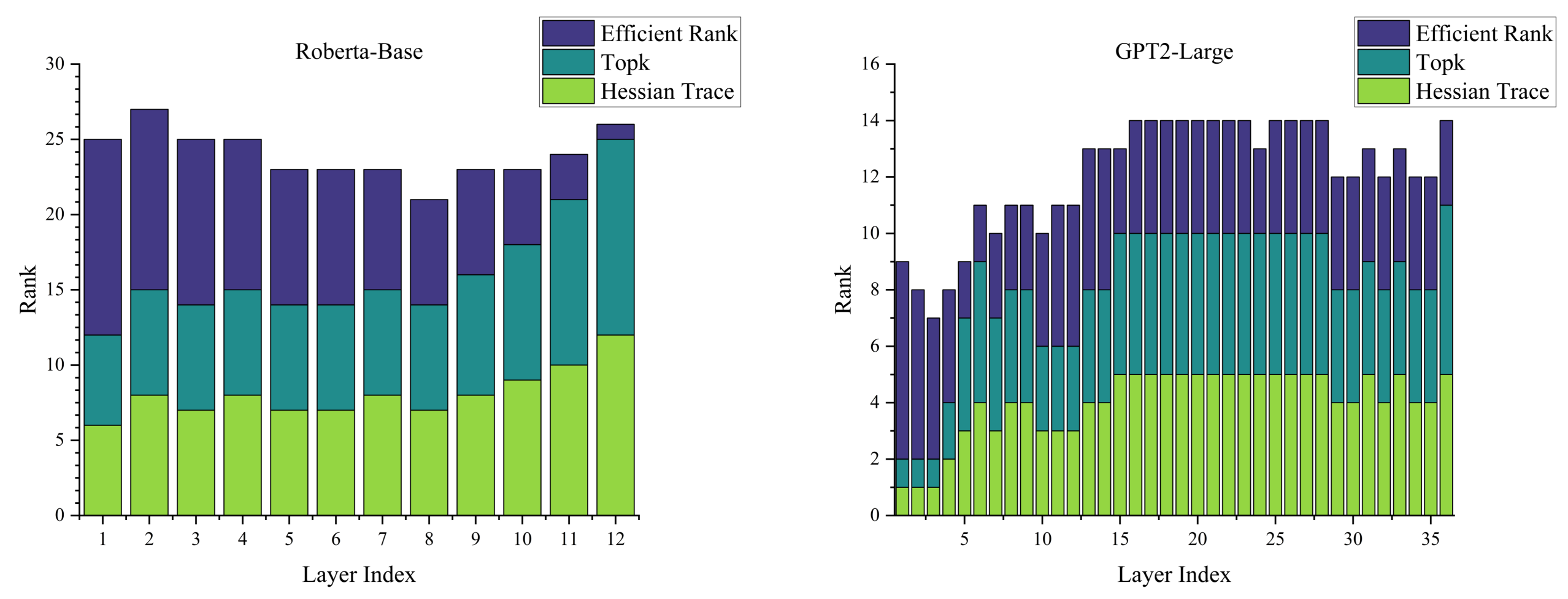}
  \vspace{-10pt}
  \caption {The rank allocation for each layer of GPT-2 Large and RoBERTa-base under different rank allocation metrics. Different colors represent different allocation metrics, and the height of each bar in the histogram corresponds to the rank allocated to that layer by the respective metric.}
    \label{rank_allocation}
\end{figure*}

\subsection{Rank Allocation Under Different Metrics}
In Figure \ref{rank_allocation}, we present the global and local rank allocation results for GPT-2 Large and RoBERTa-base, utilizing the SRA rank assignment method detailed in Section \ref{3.3}. As illustrated, the global sensitivity metric, Hessian Trace, allocates a larger rank budget to the intermediate and deeper layers of the models, with relatively less emphasis on the initial layers. In contrast, the local sensitivity metric, \textit{Topk}, primarily focuses on the middle layers, assigning more ranks to these regions. The \textit{Effective Rank} approach, however, assigns higher ranks to the initial layers and exhibits a decreasing trend in rank allocation for subsequent layers. Each of these three sensitivity metrics highlights different aspects of the models, demonstrating that relying on a single source of information for decision-making is insufficient. This highlights the necessity of Sensitivity-LoRA, which integrates these diverse information sources to achieve dynamic rank allocation.

\subsection{Overhead Analysis}
\label{Overhead Analysis}
\textbf{Cost of Obtaining the Hessian Matrix. }Computing the Hessian matrix is a complex process, especially for large models. Some methods propose processing the weight matrix by rows, allowing the Hessian matrix to be approximated through operations on activation values \cite{frantar2022gptq,li2025gptqv2}. Additionally, Cholesky decomposition is employed to enhance computational stability. In essence, we only need to perform forward inference on the model using a calibration set, and the intermediate results can be used to approximate the Hessian matrix. This significantly reduces the computational cost of the Hessian matrix. For example, when using the PIQA \cite{bisk2020piqa} dataset as the calibration set for LLaMA3.1-8B, the computation, including both metric calculation and rank allocation, can be completed in just 25.78 seconds. When using only a portion of the dataset, the computation can be finished in under 10 seconds without introducing significant errors. In contrast, other methods, such as AdaLoRA, which determine rank allocation during training, can significantly increase training time, ranging from minutes to hours. Compared to these methods, our approach introduces negligible additional computational overhead. Additionally, when we calibrate using different calibration sets, such as PIQA \cite{bisk2020piqa} and WikiText2 \cite{merity2016pointer}, we obtain nearly identical results for the Hessian matrix, which further validates the stability of our method.

\textbf{Memory Analysis. }Our method allocates the rank before training, whereas other approaches, such as AdaLoRA, require continuous rank reallocation during training. Specifically, our method has almost exactly the same memory footprint as the conventional LoRA method during training, without introducing any additional overhead. This design not only avoids the extra burden associated with dynamic rank reallocation but also ensures the efficiency and stability of the training process.

\begin{figure}[]
    \centering
    \includegraphics[width=0.9\linewidth]{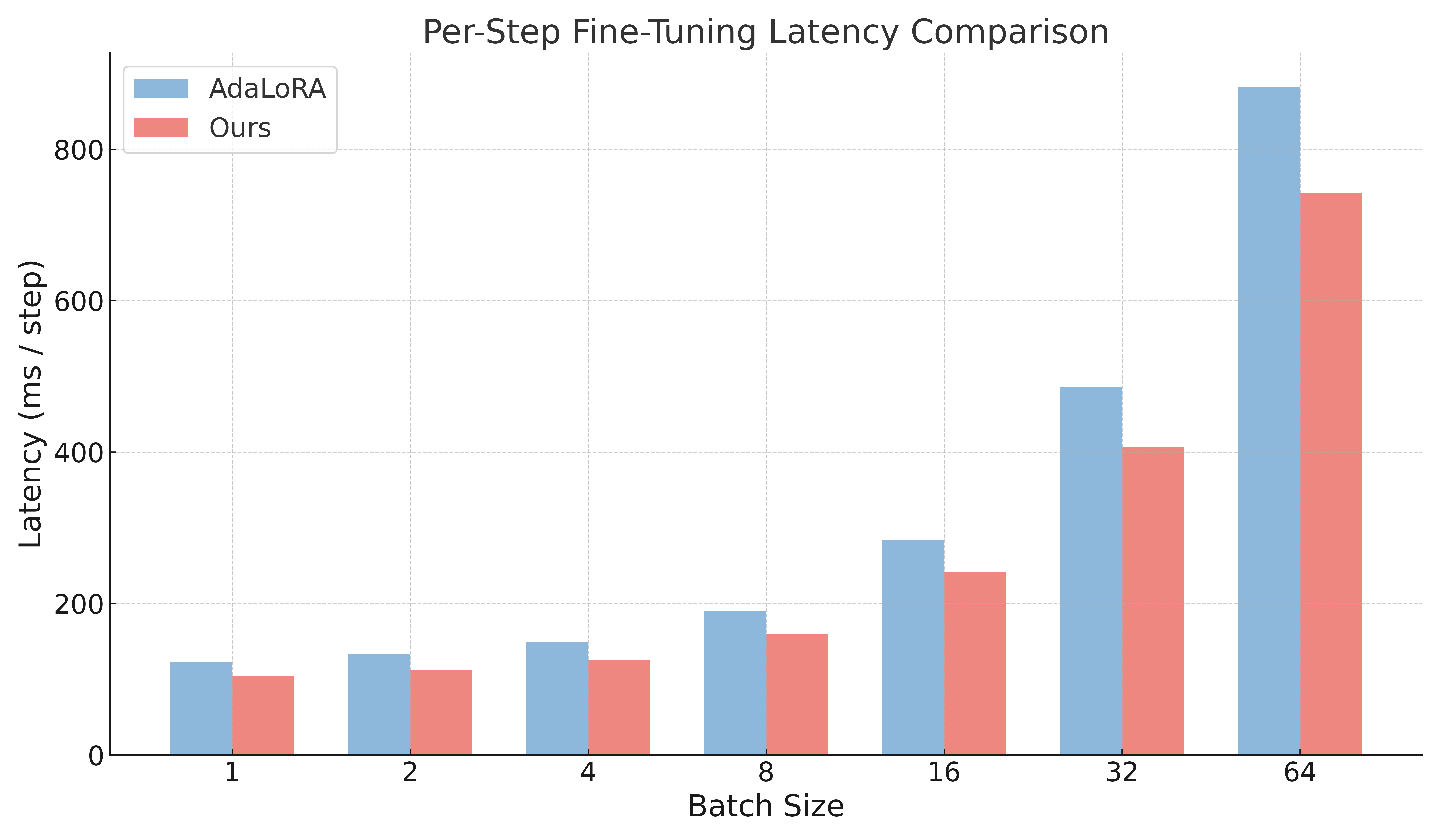}
    \vspace{-10pt}
    \caption{Comparison of per-step fine-tuning latency (ms) between AdaLoRA and ours across varying batch sizes.}
    \label{fig:runtime}
    \vspace{-10pt}
\end{figure}

\textbf{Latency Analysis. }To assess the training efficiency of our method, we measure its per-step latency and compare it with that of AdaLoRA across various batch sizes, as shown in Figure \ref{fig:runtime}. Our approach consistently exhibits lower latency under all configurations, with the performance gap widening as the batch size increases. This trend indicates superior scalability of our method. The improvement is attributed to our sensitivity-aware rank assignment strategy, which eliminates the runtime overhead associated with AdaLoRA’s dynamic scheduling. These results confirm that our method enables more efficient adaptation while significantly reducing computational costs.

\subsection{Robustness Analysis}
During fine-tuning, the calibration data can be the training data itself, making the calibration set identical in distribution to the training set, or it can consist of other non-training data. The computation of the Hessian is robust across different calibration sets and domains. To further validate this, we calibrate LLaMA3.1-8B using calibration sets from three distinct domains: Sentiment140 (sentiment analysis), PubMed 20k RCT (biomedical), and LexGLUE (legal). We denote the resulting rank orderings of the weight matrices for these domains as $r_{\text{Sentiment140}}$, $r_{\text{PubMed}}$, and $r_{\text{LexGLUE}}$, respectively. To quantify the consistency of these rankings, we employ Kendall's Tau coefficient. As shown in Table~\ref{tab:kendall}, the Kendall's Tau values between rank orders across different datasets all exceed 0.9, indicating that the weight matrix rankings are nearly identical. This confirms the strong robustness of our method across diverse domains.

\begin{table}[h]
\centering
\begin{tabular}{c|ccc}
\toprule
 & $r_{\text{Sentiment140}}$ & $r_{\text{PubMed}}$ & $r_{\text{LexGLUE}}$ \\
\midrule
$r_{\text{Sentiment140}}$ & 1.0000 & 0.9712 & 0.9834 \\
$r_{\text{PubMed}}$       & 0.9712 & 1.0000 & 0.9727 \\
$r_{\text{LexGLUE}}$      & 0.9834 & 0.9727 & 1.0000 \\
\bottomrule
\end{tabular}
\caption{Kendall's Tau correlation coefficients between rank orderings across different domains.}
\label{tab:kendall}
\end{table}

To investigate the impact of calibration set size on the rank ordering, we calibrate LLaMA3.1-8B using varying proportions of the PIQA dataset, taking the result obtained with 100\% of the data as the baseline. As shown in Table~\ref{tab:piqa_size}, using only approximately 10\% of the data already achieves a rank ordering close to that obtained with the full dataset.

\begin{table}[h]
\centering
\begin{tabular}{cc}
\toprule
Calibration Set Size & Kendall's Tau \\
\midrule
10\%  & 0.9862 \\
20\%  & 0.9876 \\
50\%  & 0.9981 \\
100\% & 1 \\
\bottomrule
\end{tabular}
\caption{Effect of calibration set size on rank ordering. "100\%" denotes the full calibration set and serves as the baseline.}
\label{tab:piqa_size}
\end{table}

We fine-tune LLaMA3.1-8B on the MNLI dataset for 5 epochs. After each epoch, we remeasure the rank ordering. Using the initially obtained rank order as the reference, we compute the Kendall's Tau coefficient between it and the rank order measured after each epoch. As shown in Table~\ref{tab:mnli_dynamic}, the Kendall's Tau values remain above 0.95 throughout training, indicating a high consistency between the dynamic rank ordering during fine-tuning and the initial static allocation.
\begin{table}[h]
\centering
\resizebox{0.45\textwidth}{!}{%
\begin{tabular}{c|cccccc}
\toprule
Epoch & 0 & 1 & 2 & 3 & 4 & 5 \\
\midrule
Kendall's Tau & 1 & 0.99 & 0.99 & 0.98 & 0.99 & 0.98 \\
\bottomrule
\end{tabular}%
}
\caption{Kendall's Tau of rank orderings measured after each epoch, using the initial static allocation (epoch 0) as the reference.}
\label{tab:mnli_dynamic}
\end{table}

\subsection{Additional Results}
We conduct experiments to validate the effectiveness of the design of the allocation parameters ($\beta_1, \beta_2, \gamma_1, \gamma_2$). The results demonstrate that combining standard deviation with scale normalization can achieve more effective rank allocation (Appendix \ref{Effectiveness of Allocation Parameter}). Additionally, we investigate the impact of hyperparameters $k$ and $\alpha$ on the performance of our method. The experiments show that our approach maintains robust performance across various hyperparameter configurations (Appendix \ref{Hyperparameter Analysis}). Furthermore, we test our method on specific text examples and obtain favorable results (Appendix \ref{Case Study}).

\section{Conclusion}
In this work, we introduce Sensitivity-LoRA, a method that efficiently allocates ranks to weight matrices based on their sensitivity, without a significant computational burden. 
Sensitivity-LoRA first performs sensitivity utilization by analyzing both global and local sensitivities. It utilizes the second-order derivatives (Hessian matrix) of the loss function to accurately capture parameter sensitivity. Next, it optimizes rank allocation by aggregating global and local sensitivities, ensuring a comprehensive and fair evaluation metric.  Extensive experiments consistently demonstrate the efficiency, effectiveness and stability of our method.

\section{Limitations}
In this paper, we conduct extensive experiments on large language models to validate the effectiveness of our proposed finetuning method. While our findings demonstrate the potential of the method in enhancing model performance, there are still areas that warrant further exploration. Specifically, we do not yet extend our evaluation to large vision models and multimodal large language models, which could provide additional insights into the generalizability and scalability of our approach. Addressing these domains will be a key focus in future work. Additionally, although our method shows promising results in the tested datasets, its robustness under low-resource and domain-specific datasets, such as those involving medical or scientific data, remains to be thoroughly assessed. Exploring these datasets could reveal further nuances in the method’s adaptability and potential for specialized applications.

\bibliography{acl}



\clearpage
\appendix



\section{Experimental Setup}

\subsection{Baselines}
\label{Baselines}
We adopt several representative methods, including  HAdapter \cite{houlsby2019HAdapter}, PAdapter \cite{pfeiffer2020PAdapter}, LoRA \cite{hu2022lora} with uniform rank allocation, AdaLoRA \cite{zhang2023adalora} and DyLoRA \cite{valipour2022dylora}, as our baselines. HAdapter \cite{houlsby2019HAdapter} and PAdapter \cite{pfeiffer2020PAdapter} are parameter-efficient fine-tuning methods based on adapters. They achieve rapid adaptation to specific tasks by inserting lightweight adapter modules into pre-trained models, eliminating the need to fine-tune the entire model. LoRA \cite{hu2022lora} approximates parameter updates by adding low-rank decomposition matrices to the weight matrices of pre-trained models, thereby reducing the number of parameters required for fine-tuning. AdaLoRA \cite{zhang2023adalora} dynamically adjusts the rank of low-rank matrices in different model layers to match their varying contributions to model performance. DyLoRA \cite{valipour2022dylora} is a dynamic low-rank adaptation technique that sorts the representations learned by adapter modules at different ranks during training and trains LoRA blocks to cover a range of ranks rather than a single rank.

\subsection{Implementation Details} 
\label{sec:detailed_exp_setup}
Our code is implemented using the PyTorch \cite{paszke2019pytorch}  framework and Transformers \cite{wolf2020transformers} libraries, with all experiments conducted on four NVIDIA A100 GPUs. When we calibrate using different calibration sets, such as PIQA \cite{bisk2020piqa} and WikiText2 \cite{merity2016pointer}, we obtain nearly identical results for the Hessian matrix, which further validates the stability of our method. The details of the approximate Hessian matrix computation can be found in Section \ref{Overhead Analysis}. We designate the local metric $S_{Topk}$ with $k$ set to half of the total number of diagonal elements, and set the parameter $\alpha$ in the \textit{Effective Rank} metric to 0.85. The values of some parameters (\(\beta_1, \beta_2, \gamma_1, \gamma_2\)) follow the settings described in Section \ref{Methodology}, and the effectiveness of this design is demonstrated in Section \ref{Experiments}. We set the average rank of each matrix to 4 for NLU and 8 for NLG. The comparison methods are required to use a similar number of finetuning parameters. The training is performed using the Adam optimizer with a learning rate of \(5 \times 10^{-4}\), a batch size of 32 for 10 epochs.

\section{Parameter Analysis}
\subsection{Effectiveness of Allocation Parameter}
\label{Effectiveness of Allocation Parameter}
We conduct validation experiments on the allocation parameters  (\(\beta_1, \beta_2, \gamma_1, \gamma_2\)) to assess the effectiveness of the proposed weighted formula (which is consistently applied to the local sensitivity weight \(\beta_1, \beta_2\) and the global local fusion weight \(\gamma_1, \gamma_2\), as mentioned in Section \ref{Methodology}). Specifically, we compare our design with a method that does not consider the standard deviation of the metrics and instead assigns equal weights to each metric (using \( \beta_1 = \frac{0.5}{\mu^{S_T}},\beta_2 = \frac{0.5}{\mu^{S_E}},\gamma_1 = \frac{0.5}{\mu^{S_g}},\gamma_2 = \frac{0.5}{\mu^{S_l}} \)). As shown in Table \ref{tab:weight}, our parameter strategy achieves superior performance across all tasks, with an average score of 85.94 compared to the other of 85.65. The performance improvements are particularly significant in the RTE, CoLA and SST-2 tasks, indicating that our approach has better adaptability to different tasks. These results demonstrate that combining the standard deviation with scale normalization can achieve more expressive and stable sensitivity modeling, thereby enabling more effective rank allocation and overall finetuning performance.


\begin{table*}[htbp]
\centering
\small
\begin{tabular}{lccccccccc}
\toprule
\textbf{Method} & \textbf{MNLI} & \textbf{SST-2} & \textbf{MRPC} & \textbf{CoLA} & \textbf{QNLI} & \textbf{QQP} & \textbf{RTE} & \textbf{STS-B} & \textbf{Avg.} \\ 
\midrule

$0.5/\mu$ & 87.49 & 94.32 & 87.61 & 59.90 & 93.47 & 90.69 & 80.50 & 91.18 & 85.65 \\

\rowcolor[gray]{0.80}
Ours      & \textbf{87.58} & \textbf{94.59} & \textbf{87.73} & \textbf{60.20} & \textbf{93.62} & \textbf{90.74} & \textbf{81.81} & \textbf{91.27} & \textbf{85.94} \\ 

\bottomrule
\end{tabular}
\caption{Performance comparison of our allocation weight parameter strategy and equal allocation weight parameter. We compare our proposed formulation \(\beta_1, \beta_2, \gamma_1, \gamma_2\) (\(\sigma/\mu^{2}\))
with a baseline \(\beta_1, \beta_2, \gamma_1, \gamma_2\) (\(0.5/\mu\)). Results are reported on the GLUE benchmark using RoBERTa-base.}
\label{tab:weight}
\end{table*}

\subsection{Hyperparameter Analysis}
\label{Hyperparameter Analysis}
In this study, we delve into the influence of the hyperparameters $k$ and $\alpha$ on the performance of our method. These hyperparameters are crucial for the computation of the \textit{Topk} and \textit{Effective Rank} metrics, respectively. As illustrated in Table \ref{tab:k_alpha}, we examine two representative configurations: $k = \frac{N}{3}, \alpha = 0.80$ and $k = \frac{N}{2}, \alpha = 0.85$, where $N$ denotes the total number of diagonal elements in the Hessian matrix. Under both settings, our method demonstrates remarkable consistency, achieving average scores of 85.93 and 85.94, respectively. These strong results highlight the robustness of our sensitivity based rank allocation framework to reasonable variations in the metric configuration. Moreover, they underscore the effectiveness of integrating the \textit{Topk} and \textit{Effective Rank} metrics, which successfully capture salient parameter sensitivities across different thresholds.


\begin{table*}[htbp]
\centering
\small

\begin{tabular}{cc|ccccccccc}
\toprule
\textbf{$k$} & \textbf{$\alpha$} & \textbf{MNLI} & \textbf{SST-2} & \textbf{MRPC} & \textbf{CoLA} & \textbf{QNLI} & \textbf{QQP} & \textbf{RTE} & \textbf{STS-B} & \textbf{Avg.} \\ 
\midrule

$N/3$ & 0.80 & 87.55 & 94.52 & 87.72 & \textbf{60.21} & 93.62 & 90.70 & \textbf{81.82} & \textbf{91.34} & 85.93 \\

$N/2$ & 0.85 & \textbf{87.58} & \textbf{94.59} & \textbf{87.73} & {60.20} & \textbf{93.62} & \textbf{90.74} & {81.81} & {91.27} & \textbf{85.94} \\ 

\bottomrule
\end{tabular}
\caption{Performance comparison of the RoBERTa-base under different hyperparameters ($k$ and $\alpha$). The $k$ denotes the number of top Hessian diagonal elements used in $S^{w}_{{Topk}}$, and $\alpha$ is the cumulative contribution threshold used in $S^{w}_{{EffectiveRank}}$.}
\label{tab:k_alpha}
\end{table*}

\section{Case Study}
\label{Case Study}

\begin{figure}[t]
    \centering
    \includegraphics[width=1\linewidth]{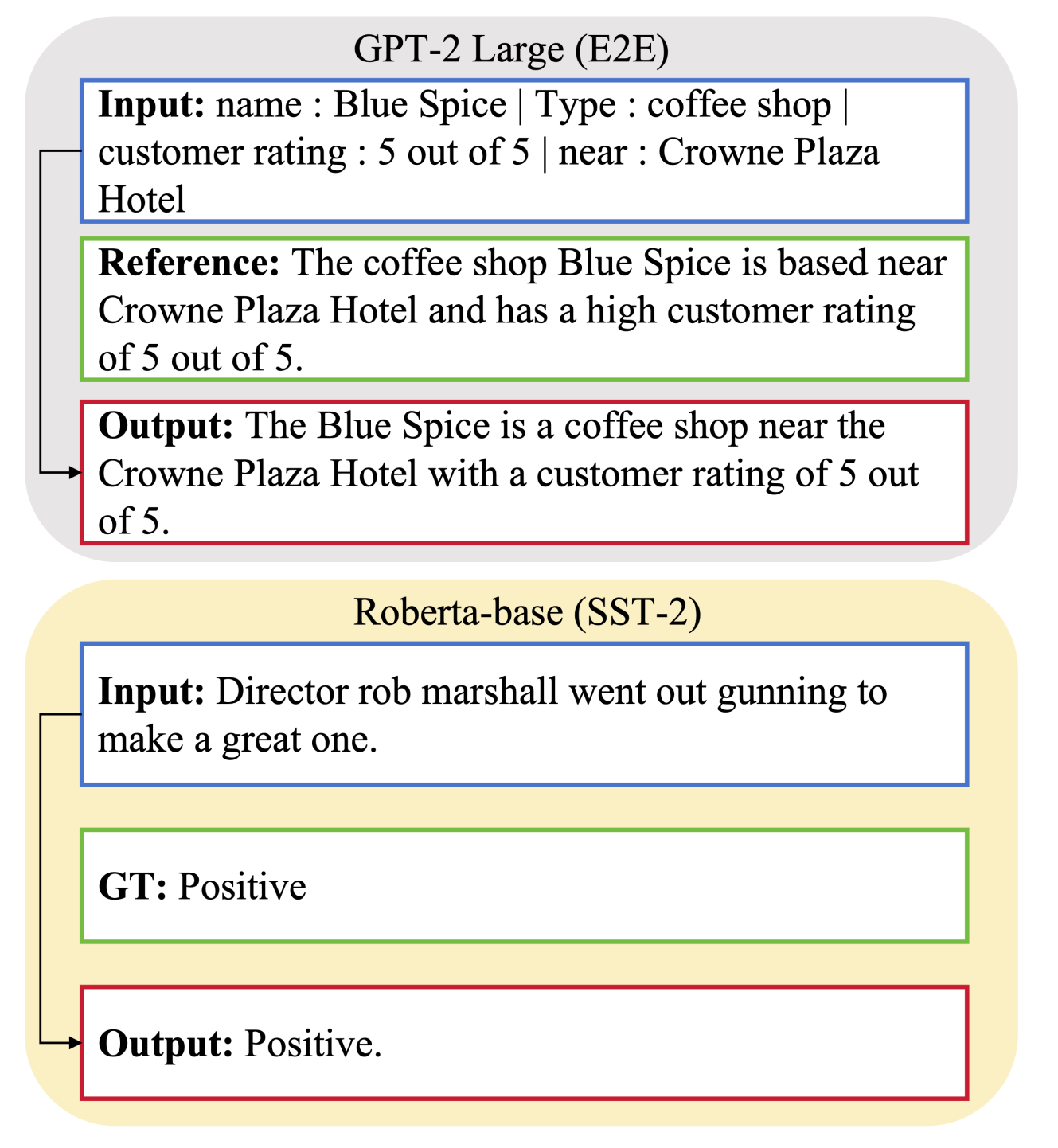}
    \caption{The Case Study of the GPT-2 Large and RoBERTa-base models. The blue boxes represent the input test data, the green boxes indicate the reference text or ground truth output, and the red boxes highlight the model's actual output.}
    \label{fig:case_study}
\end{figure}

Figure \ref{fig:case_study} presents the performance of the GPT-2 Large and RoBERTa-base models fine-tuned using the dynamic rank allocation method (Sensitivity-LoRA) on the E2E and SST-2 datasets. For the E2E dataset, the GPT-2 Large model generates fluent and grammatically correct natural language text that closely aligns with the reference while retaining the input information. This indicates that the model effectively processes structured inputs and excels at generating accurate and coherent natural language descriptions. For the SST-2 dataset, the RoBERTa-base model achieves strong performance in sentiment classification tasks, accurately classifying input text as "Positive." These results demonstrate the effectiveness of the Sensitivity-LoRA method in enhancing model performance on both text generation and classification tasks.

\section{Analysis on Scalability and Generality}
We conduct additional experiments to demonstrate that our method remains efficient on large-scale models. Table~\ref{tab:qwen_time} compares our method with {AdaLoRA} on {Qwen2.5-32B} (MNLI dataset, 8$\times$H100 GPUs), including preprocessing, training, and total time. Table~\ref{tab:qwen_perf} shows that our method consistently achieves the best performance.

\begin{table}[htbp]
\centering
\begin{tabular}{c|ccc}
\toprule
Method & Preprocess & Train& Total\\
\midrule
AdaLoRA & 0   & 9725 & 9725 \\
Ours    & 178 & 6897 & 7075 \\
\bottomrule
\end{tabular}
\caption{Time (s) comparison of fine-tuning methods with different rank allocations on {Qwen2.5-32B}.}
\label{tab:qwen_time}
\end{table}

\begin{table}[htbp]
\centering
\begin{tabular}{c|cc}
\toprule
Method & GPQA & HumanEval \\
\midrule
AdaLoRA & 46.21 & 55.58 \\
DyLoRA  & 45.84 & 56.04 \\
Ours    & 46.66 & 56.88 \\
\bottomrule
\end{tabular}
\caption{Fine-tuning performance of {Qwen2.5-32B} on GPQA and HumanEval.}
\label{tab:qwen_perf}
\end{table}

Moreover, we evaluate our method on the COCO2017 dataset using the {LLaVA1.5-7B} model, with CIDEr and ROUGE\_L as evaluation metrics. As shown in Table~\ref{tab:llava}, our approach outperforms existing methods on multimodal tasks.

\begin{table}[htbp]
\centering
\begin{tabular}{c|cc}
\toprule
Method & CIDEr & ROUGE\_L \\
\midrule
AdaLoRA & 1.0284 & 0.5874 \\
DyLoRA  & 1.0175 & 0.5856 \\
Ours    & 1.0561 & 0.5950 \\
\bottomrule
\end{tabular}
\caption{Comparison of different fine-tuning methods on COCO2017 with \textsc{LLaVA1.5-7B}.}
\label{tab:llava}
\end{table}

We re-evaluate the {RoBERTa-base} model on the {MNLI} dataset using 8$\times$H100 GPUs with a batch size of 8. As shown in Table~\ref{tab:roberta_mnli}, our method demonstrates clear advantages over the standard AdaLoRA and LoRA in both memory usage and total fine-tuning time. While LoRA is slightly faster than our method in terms of per-epoch training time, our method converges to the optimal solution significantly faster overall.

\begin{table}[h]
\centering
\resizebox{0.45\textwidth}{!}{
\begin{tabular}{c|ccc}
\toprule
Method &Memory (MB) &Epoch (s) & Total (s) \\
\midrule
LoRA    & 3899 & 368  & 1508 \\
AdaLoRA & 4115 & 425  & 2981 \\
\textbf{Ours} & \textbf{3887} & 371 & \textbf{1371} \\
\bottomrule
\end{tabular}
}
\caption{Memory usage, single epoch latency, and total convergence time under different methods.}
\label{tab:roberta_mnli}
\end{table}

We include a full-parameter (Full) baseline under the same experimental setup. As shown in Table~\ref{tab:glue_full}, our method achieves performance comparable to the Full model, with only marginal differences of less than 1\%.

\begin{table}[h]
\centering
\begin{tabular}{c|cccc}
\toprule
Method & MNLI & SST-2 & MRPC & CoLA \\
\midrule
Full & 87.63 & 94.69 & 87.73 & 60.28 \\
Ours & 87.58 & 94.59 & 87.73 & 60.20 \\
\bottomrule
\end{tabular}
\caption{Comparison between our method and full-parameter fine-tuning on the GLUE benchmark with {RoBERTa-base}.}
\label{tab:glue_full}
\end{table}

\section{Details of Efficient Hessian Matrix Approximation}
Our Hessian approximation uses the autocorrelation matrix of activations as a surrogate for second-order sensitivity, efficiently capturing the impact of parameter updates on model performance.In practice, we use a block-wise strategy to partition the large-scale Hessian matrix into smaller sub-blocks, enabling efficient factorization and inversion via Cholesky decomposition. This approach significantly reduces computational complexity while maintaining stability and accuracy in the error compensation process.For layers with an extremely large number of parameters, the difference between strict per-row optimization order and a uniform quantization order is negligible. Consequently, a consistent order is adopted across all rows, ensuring that the Hessian matrices corresponding to each row are identical and derived solely from shared input activations rather than weight-specific variations. This design choice allows the sensitivity updates related to parameter changes to be computed once and shared across all rows, avoiding redundant computations inherent in row-specific update paths. As a result, the unified order substantially improves computational efficiency and memory utilization.

\end{document}